\documentclass{article}

\PassOptionsToPackage{numbers, compress}{natbib}
\usepackage[final]{nips_2016}

\usepackage[utf8]{inputenc} 
\usepackage[T1]{fontenc}    
\usepackage{hyperref}       
\usepackage{url}            
\usepackage{booktabs}       
\usepackage{amsfonts}       
\usepackage{nicefrac}       
\usepackage{microtype}      

\usepackage[colorinlistoftodos]{todonotes}

\usepackage{xspace}
\newcommand*{\ie}{i.e.\@\xspace}

\usepackage{graphicx}

\usepackage{listings}
\usepackage{color}

\usepackage[scaled]{beramono}
\definecolor{pythonfile}{rgb}{0,0.6,0.8}
\definecolor{codegreen}{rgb}{0,0.6,0}
\definecolor{codegray}{rgb}{0.5,0.5,0.5}
\definecolor{codepurple}{rgb}{0.58,0,0.82}
\definecolor{backcolour}{rgb}{0.95,0.95,0.92}
 
\lstdefinestyle{mystyle}{
    backgroundcolor=\color{backcolour},   
    commentstyle=\color{codegreen},
    keywordstyle=\color{magenta},
    numberstyle=\tiny\color{codegray},
    stringstyle=\color{codepurple},
    basicstyle=\ttfamily\scriptsize,
    breakatwhitespace=false,         
    breaklines=true,                 
    captionpos=b,                    
    keepspaces=true,                 
    numbers=left,                    
    numbersep=5pt,                  
    showspaces=false,                
    showstringspaces=false,
    showtabs=false,                  
    tabsize=2
}
\title{A User Simulator for Task-Completion Dialogues\thanks{The source code is available at: \url{https://github.com/MiuLab/UserSimulator}}}

\author{Xiujun Li$^\dag$\quad Zachary C. Lipton$^\star$\quad Bhuwan Dhingra$^{\ddag}$\quad \\ \textbf{Lihong Li$^\dag$}\quad \textbf{Jianfeng Gao$^\dag$}\quad \textbf{Yun-Nung Chen}$^{\S}$\quad \\
  $^\dag$Microsoft Research, Redmond, WA, USA \\
  $^\star$University of California, San Diego, CA, USA \\
  $^\ddag$Carnegie Mellon University, Pittsburgh, PA, USA \\
  $^\S$National Taiwan University, Taipei, Taiwan \\
  {\small \tt $^\dag$\{xiul,lihongli,jfgao\}@microsoft.com $^\star$zlipton@cs.ucsd.edu}\\ 
  {\small \tt $^\ddag$bdhingra@andrew.cmu.edu $^\S$y.v.chen@ieee.org}
 }

\begin{document}

\maketitle

\begin{abstract}
Despite widespread interests in reinforcement-learning  for task-oriented dialogue systems, 
several obstacles can frustrate research and development progress. 
First, reinforcement learners typically require interaction with the environment, 
so conventional dialogue corpora cannot be used directly. 
Second, each task presents specific challenges, 
requiring separate corpus of task-specific annotated data. 
Third, collecting and annotating human-machine or human-human conversations 
for task-oriented dialogues requires extensive domain knowledge. 
Because building an appropriate dataset can be both financially costly and time-consuming, 
one popular approach is to build a user simulator based upon a corpus of example dialogues. 
Then, one can train reinforcement learning agents in an online fashion 
as they interact with the simulator. 
Dialogue agents trained on these simulators 
can serve as an effective starting point. 
Once agents master the simulator, 
they may be deployed in a real environment to interact with humans, 
and continue to be trained online. 
To ease empirical algorithmic comparisons in dialogues, 
this paper introduces a new, publicly available simulation framework, 
where our simulator, designed for the movie-booking domain, 
leverages both rules and collected data. 
The simulator supports two tasks: movie ticket booking and movie seeking. 
Finally, we demonstrate several agents 
and detail the procedure to add and test your own agent in the proposed framework.
\end{abstract}

\section{Introduction}
\label{sec:introduction}
Practical dialogue systems consist of several components. The \emph{natural language understanding} (NLU) module maps free texts to structured semantic frames of utterances. The \emph{natural language generation} (NLG) module maps the structured representations back into a natural-language form. \emph{Knowledge bases} (KBs) and \emph{state trackers} provide access to side information and track the evolving state of the dialogue, respectively. The \emph{dialogue policy} is a central component of the system that chooses an \emph{action} given the current state of the dialogue. 

In traditional systems, dialogue policies might be programmed explicitly with rules. However, rule-based approaches have several weaknesses. First, for complex systems, it may not be easy to design a reasonable rule-based policy. Second, the optimal policy might change over time, as user behavior changes. A rule-based system cannot cope with such non-stationarity. Thus, reinforcement learning, in which policies are learned automatically from experience, offers an appealing alternative.

\subsection{Why Is User Simulation Needed?}
%
%
Typically, researchers seek to optimize dialogue policies with either supervised learning (SL) or reinforcement learning (RL) methods. In SL approaches, a policy is trained to imitate the observed actions of an expert. Supervised learning approaches often require a large amount of expert-labeled data for training. For task-specific domains, intensive domain knowledge is usually required for collecting and annotating actual human-human or human-machine conversations, and is often expensive and time-consuming. Additionally, even with a large amount of training data, it is possible that some dialogue state spaces may not be explored sufficiently in the training data, preventing a supervised learner to find a good policy.

In contrast, RL approaches allow an agent to learn without any expert-generated example. Given only a reward signal, the agent can optimize a dialogue policy through interaction with users. Unfortunately, RL can require many samples from an environment, making learning from scratch with real users impractical. To overcome this limitation, many researchers in the dialogue systems community train RL agents using simulated users ~\cite{cuayahuitl2005human,eckert1997user,georgila2005learning,pietquin2006consistent,pietquin2006probabilistic,schatzmann2006survey,scheffler2002automatic}. 

The goal of user simulation is to generate natural and reasonable conversations, allowing the RL agent to explore the policy space. The simulation-based approach allows an agent to explore trajectories which may not exist in previously observed data, overcoming a central limitation of imitation-based approaches. Dialogue agents trained on these simulators can then serve as an effective starting point, after which they can be deployed against real humans to improve further via reinforcement learning. 

\subsection{Related Work}
%
%
\label{sec:related_work}
%
%
Given the reliance of the research community on user simulations, it seems important to assess the quality of the simulator. How best to assess a user simulator remains an open issue, and there is no universally accepted metric~\cite{pietquin2013survey}. 
One important feature of a good user simulator requires coherent behavior throughout the dialogue; ideally, a good metric should measure the correlation between user simulation and real human behaviors, but it is hard to find a widely accepted metric.  Therefore, to the best of our knowledge, there is no standard way to build a user simulator. Here, we summarize the literature of user simulation in different aspects:

\begin{itemize}
\item{At the granularity level, the user simulator can operate either at the \emph{dialog-act}\footnote{Here, a dialog-act consists of one intent, as well as zero, one or multiple slot-value pairs.  In the rest of the paper, we will use dialog-acts and dialog actions interchangeably} level, 
or at the \emph{utterance} level~\cite{jung2009data}.}
\item{At the methodology level, the user simulator could use a \emph{rule-based} approach, or a \emph{model-based} approach where the model is learned from training data.}
\end{itemize}

Many models have been introduced for user modeling in different dialogue systems. Early work~\cite{eckert1997user,levin2000stochastic} employed a simple, naive bi-gram model $P(a_u|a_m)$ to predict the next user-act $a_u$ based on the last system-act $a_m$. The parameters of this model are simple, but it cannot produce coherent user behaviors, for two reasons: (1) this model can only look at the last system action, and (2) if the user changes its goal, this bi-gram model might produce some illogical behavior since it does not consider the user goal when generating the next user-act. Much of the follow-up work on user simulators has tried to address these issues. The first issue can be addressed by looking at longer dialogue histories to select the next user action~\cite{frampton2006learning,georgila2005learning}; the second issue can be attacked by explicitly incorporating the user goal into user state modeling~\cite{scheffler2003automatic}. 

The recently proposed sequence-to-sequence approach~\cite{sutskever2014sequence} has inspired end-to-end trainable user simulators~\cite{asri2016sequence}. This approach treats user-turn dialogue to agent-turn dialogue as a source-to-target sequence generation problem, which might be suitable for chatbot-like systems, but may not work well for domain-specific, task-completion dialogue systems, which require the ability to interact with databases and aggregate useful information into the system responses. The benefit of such model-based approaches is they do not need intensive feature engineering, but they typically require a large amount of labeled data to generalize well and deal with user states not included in the training data. On the other hand, agenda-based user simulation~\cite{schatzmann2009hidden} provides a convenient mechanism to explicitly encode the dialogue history and user goal. The user goal consists of slot-value pairs describing the user's requests and constraints. A stack-like format models the state transitions and user action generation as a sequence of simple push and pop operations, which ensures the consistency of user behavior over the course of conversation. 

In this paper, we combine the benefits of both model-based and rule-based approaches. Our user simulation for the task-completion dialogue setting follows an agenda-based approach at the dialog-act level, and a sequence-to-sequence natural language generation (NLG) component is used to convert the selected dialog-act into natural language.

\section{Dialogue Systems for Task-Completion}
\label{sec:tcdialog}

We consider a dialogue system for helping users to book movie tickets or to look up the movies they want, by interacting with them in natural language. Over the course of conversation, the agent gathers information about the customer’s desires and ultimately books the movie tickets, or identify the movie of interest. The environment then assesses a binary outcome (success or failure) at the end of the conversation, based on (1) whether a movie is booked, and (2) whether the movie satisfies the user’s constraints.

\paragraph{Data:} The data we used in the paper was collected via Amazon Mechanical Turk, and the annotation was done internally using our own schema. There are $11$ intents (\ie, inform, request, confirm\_question, confirm\_answer, etc.), and $29$ slots (\ie, moviename, starttime, theater, numberofpeople, etc.). Most of the slots are \textit{informable} slots, which users can use to constrain the search, and some are \textit{requestable} slots, of which users can ask values from the agent. For example, \emph{numberofpeople} cannot be a requestable slot, since arguably user knows how many tickets he or she wants to buy. In total, we labeled $280$ dialogues in the movie domain, and the average number of turns per dialogue is approximately $11$.

\section{User Simulator}
\label{sec:usersim}
In this work, we follow the agenda-based user simulation approach~\cite{schatzmann2009hidden}, in which a stack-like representation of user state provides a convenient mechanism to explicitly encode the dialogue history and user's goal, and user state update (state transition and user action generation) can be modeled as sequences of push and pop operations with stacks. Here, we describe the rule-based user simulator in detail.

\subsection{User Goal} 
In the task-oriented dialogue setting, the first step of user simulation is to generate a user goal; the agent knows nothing about the user goal but its objective is to help the user to accomplish this goal. Hence, the entire conversation exchange is around this goal implicitly. Generally, the definition of user goal contains two parts: 

\begin{itemize}
\item \emph{inform\_slots} contain a number of slot value pairs which serve as constraints from the user.
\item \emph{request\_slots} contain a set of slots that user has no information about the values, but wants to get the values from the agent side during the conversation.
\end{itemize}

To make the user goal more realistic, we add some constraints in the user goal: Slots are split into two groups. For movie-booking scenario, some of elements must appear in the user goal, we called these elements as \emph{Required slots}, which includes \emph{moviename, theater, starttime, date, numberofpeople}; the rest slots are \emph{Optional slots}; \emph{ticket} is a default slot which always appears in the \emph{request\_slots} part of user goal.

We generated the user goals from the labeled dataset, using two mechanisms. One mechanism is to extract all the slots (known and unknown) from the first user turns (excluding the greeting user turn) in the data, since usually the first turn contains some or all the required information from user. The other mechanism is to extract all the slots (known and unknown) that first appear in all the user turns, and then aggregate them into one user goal. We dump these user goals into a file as the user-goal database for the simulator. Every time when running a dialogue, we randomly sample one user goal from this user goal database.

\subsection{User Action}

\paragraph{First user-act:} 
The work focuses on user-initiated dialogues, so we randomly generated a user goal as the first turn (a user turn).
To make the user-act more reasonable, we add further constraints in the generation process.  For example, the first user turn is usually a request turn; it has at least one informable slot; if the user knows the movie name, \emph{moviename} will appear in the first user turn; etc.

During the course of a dialogue, the user simulator maintains a compact stack-like representation named as \emph{user agenda}~\cite{schatzmann2009hidden}, where the user state $s_u$ is factored into an agenda $A$ and a goal $G$, which consists of constraints $C$ and request $R$. At each time-step $t$, the user simulator will generate the next user action $a_{u,t}$ based on the its current status $s_{u,t}$ and the last agent action $a_{m,t-1}$, and then update the current status $s'_{u,t}$. Here, when training or testing a policy without natural language understanding (NLU), an error model~\cite{schatzmann2007error} is introduced to simulate the noise from the NLU component, and noisy communication between the user and agent. There are two types of noise channels in the error model: one is at the intent level, the other is slot level. Furthermore, at the slot level, there are three kinds of possible noise:

\begin{itemize}
\item \emph{slot deletion}: to simulate the scenario that the slot was not recognized by the NLU;
\item \emph{incorrect slot value}: to simulate the scenario that the slot name was recognized correctly, but the slot value was not recognized correctly, e.g., wrong word segmentation;
\item \emph{incorrect slot}: to simulate the scenario that both the slot and its value were not recognized correctly.
\end{itemize}

When training or testing a policy with natural language understanding (NLU), it is not necessary to use the error model because the NLU component itself introduces noise.

If the agent action is \emph{inform(taskcomplete)}, this is to inform that the agent has gathered all the information and is ready to book the movie ticket. The user simulator will check whether the current stack is empty, and also conduct constraint checking to make sure that the agent is trying to book the right movie tickets. This guarantees that the user behaves in a consistent, goal-oriented manner.

\subsection{Dialogue Status}
There are three statuses for a dialogue: \emph{no\_outcome\_yet}, \emph{success} and \emph{failure}.  The status is \emph{no\_outcome\_yet} if the agent has not issued the \emph{inform(taskcomplete)} action and if the number of turns of the conversation has not exceeded the maximum value; otherwise, the dialogue is finished with either a \emph{success} or a \emph{failure} outcome.  To be a \emph{success} dialogue, the agent must answer all the questions (a.k.a. requestable slots of the user) and book the right movie tickets finally, within the maximum number of turns.  All other cases are \emph{failure} dialogues.  For example, the whole dialogue exceeds the limit of max turns, or the agent books the wrong movie tickets for the user.

There is a special case, where the user's constraints are not satisfiable in our movie database, and the agent correctly informs that no ticket can be booked.  One can argue this is a successful outcome, as the agent does what is correct.  Here, we choose to treat it as a failure, as no ticket is booked.  It should be noted that this choice does not affect algorithm comparison much.

\subsection{Natural Language Understanding (NLU)}
The natural language understanding (NLU) component is a recurrent neural network model with long-short term memory (LSTM) cells.
This single NLU model~\cite{hakkani2016multi} can do intent prediction, and slot filling simultaneously. For joint modeling of intent and slots, the predicted tag set is a concatenated set of IOB-format slot tags and intent tags, and an additional token \emph{<EOS>} is introduced at the end of each utterance, its supervised label is an intent tag, while the supervised label of all other preceding words is an IOB tag. In this way, we can still use the sequence-to-sequence training approach, the last hidden layer of the sequence is supposed to be a condensed semantic representation of the whole input utterance, so that it can be utilized for intent prediction at the utterance level. This model is trained using all available dialogue actions and utterance pairs in our labeled dataset.

\subsection{Natural Language Generation (NLG)}
The user simulator is designed on dialog act level, but it can also work on utterance level, we provide a natural language generation (NLG) component in the framework. Due to the limited labeled dataset, our empirical tests found that a pure model-based NLG might not generalize well, which will introduce a lot of noise for the policy training. Thus, we use a hybrid approach which consists of:

\begin{itemize}
\item \emph{Template-based NLG}: outputs some predefined rule-based templates for dialog acts
\item \emph{Model-based NLG}: is trained on our labeled dataset in a sequence-to-sequence fashion. It takes dialog-acts as input, and generates template-like sentences with slot placeholders via an LSTM decoder. Then, a post-processing scan is performed to replace the slot placeholders with their actual values~\cite{wen2015semantically,wen2016snapshot}.
In the LSTM decoder, we apply beam search, which iteratively considers the top $k$ best sentences up to time step $t$ when generating the token of the time step $t+1$. For the sake of the trade-off between the speed and performance, we use the beam size of $3$ in our experiments.
\end{itemize}

In our hybrid model, if the dialog act can be found in the predefined rule-based templates, we use the template-based NLG for generating the utterance; otherwise, the utterance is generated by the model-based NLG.

\section{Usages}
\label{sec:usages}

We conduct experiments training agents with our user simulator for the following two tasks. The first is a task-completion dialogue setting on the movie-booking domain ~\cite{lipton2016efficient}. Here, the agent's job is to engage with the user in a dialogue with the ultimate goal of helping the user to successfully book a movie. To measure the quality of the agent, there are three metrics: \{\emph{success rate\footnote{\emph{Success rate} is sometimes known as \emph{task completion rate} --- the fraction of dialoges that finish successfully.}, average reward, average turns}\}; each of them provides different information about the quality of agents. There exists a strong correlation among them: generally, a good policy should have a higher success rate, higher average reward and lower average turns. Here, we choose \emph{success rate} as our major evaluation metric to report for the quality of agents.
In the appendix, Table~\ref{tab:tcp_sample} demonstrates some example dialogues for this task. 

The second task pertains to training an KB-InfoBot~\cite{dhingra2016end}. The setting is a simplified version of the previous goal-oriented dialogues, in which an agent and user communicate with only two intents (\emph{request} and \emph{inform}). Accordingly, for this task the experiments in KB-InfoBot~\cite{dhingra2016end} engage a simplified version of the simulator described in this paper, using the two aforementioned intents and six slots. In this paper, the knowledge-base is drawn from the IMDB dataset. In the appendix, Table~\ref{tab:sample} demonstrates some example dialogues for KB-InfoBot. 

\section{Discussion}
In this paper, we demonstrated that rule-based user simulation can be a safe way to train reinforcement learning agents for task-completion dialogues.  Since rule-based user simulation requires application-specific domain knowledge to curate these hand-crafted rules, it is usually a time-consuming process. One improvement for the current user simulation in the task-completion dialogue setting is to include user goal changes which make the dialogue more complex, but also realistic. Another potential direction for future improvement is model-based user simulation for task-completion dialogues.  The advantage of model-based user simulation is that it can be adapted to other domains easily as long as there are enough labeled data. Since model-based user simulation is data-driven, one potential risk is that it asks for a large amount of labeled data to train a good simulator, and it might be risky to use the user simulator to train RL agents due to the uncertainty of the model. When training reinforcement learning agents with such a user simulator, the RL agents can easily learn these errors or loopholes existing in the model-based user simulator and make the false dialogues ``success''.  In this case, the quality of learned RL policy can be misleadingly high. But model-based user simulator for task-completion dialogue setting is still a good direction to investigate.

\section{Acknowledgments}
We thank Asli Celikyilmaz, Alex Marin, Paul Crook, Dilek Hakkani-T\"{u}r, Hisami Suzuki, Ricky Loynd and Li Deng for their insightful comments and discussion in the project.

\bibliographystyle{plain}
\bibliography{nips_2016}

\begin{thebibliography}{10}

\bibitem{asri2016sequence}
Layla~El Asri, Jing He, and Kaheer Suleman.
\newblock A sequence-to-sequence model for user simulation in spoken dialogue
  systems.
\newblock {\em arXiv:1607.00070}, 2016.

\bibitem{cuayahuitl2005human}
Heriberto Cuay{\'a}huitl, Steve Renals, Oliver Lemon, and Hiroshi Shimodaira.
\newblock Human-computer dialogue simulation using hidden markov models.
\newblock In {\em IEEE Workshop on Automatic Speech Recognition and
  Understanding}. IEEE, 2005.

\bibitem{dhingra2016end}
Bhuwan Dhingra, Lihong Li, Xiujun Li, Jianfeng Gao, Yun-Nung Chen, Faisal
  Ahmed, and Li~Deng.
\newblock End-to-end reinforcement learning of dialogue agents for information
  access.
\newblock {\em arXiv:1609.00777}, 2016.

\bibitem{eckert1997user}
Wieland Eckert, Esther Levin, and Roberto Pieraccini.
\newblock User modeling for spoken dialogue system evaluation.
\newblock In {\em Automatic Speech Recognition and Understanding, 1997.
  Proceedings., 1997 IEEE Workshop on}, pages 80--87. IEEE, 1997.

\bibitem{frampton2006learning}
Matthew Frampton and Oliver Lemon.
\newblock Learning more effective dialogue strategies using limited dialogue
  move features.
\newblock In {\em ACL}. Association for Computational Linguistics, 2006.

\bibitem{georgila2005learning}
Kallirroi Georgila, James Henderson, and Oliver Lemon.
\newblock Learning user simulations for information state update dialogue
  systems.
\newblock In {\em INTERSPEECH}, pages 893--896, 2005.

\bibitem{hakkani2016multi}
Dilek Hakkani-T{\"u}r, Gokhan Tur, Asli Celikyilmaz, Yun-Nung Chen, Jianfeng
  Gao, Li~Deng, and Ye-Yi Wang.
\newblock Multi-domain joint semantic frame parsing using bi-directional
  rnn-lstm.
\newblock In {\em Interspeech}, 2016.

\bibitem{jung2009data}
Sangkeun Jung, Cheongjae Lee, Kyungduk Kim, Minwoo Jeong, and Gary~Geunbae Lee.
\newblock Data-driven user simulation for automated evaluation of spoken dialog
  systems.
\newblock {\em Computer Speech \& Language}, 23(4):479--509, 2009.

\bibitem{levin2000stochastic}
Esther Levin, Roberto Pieraccini, and Wieland Eckert.
\newblock A stochastic model of human-machine interaction for learning dialog
  strategies.
\newblock {\em IEEE Transactions on speech and audio processing}, 8(1):11--23,
  2000.

\bibitem{lipton2016efficient}
Zachary~C Lipton, Jianfeng Gao, Lihong Li, Xiujun Li, Faisal Ahmed, and
  Li~Deng.
\newblock Efficient exploration for dialogue policy learning with {BBQ}
  networks \& replay buffer spiking.
\newblock {\em arXiv:1608.05081}, 2016.

\bibitem{pietquin2006consistent}
Olivier Pietquin.
\newblock Consistent goal-directed user model for realisitc man-machine
  task-oriented spoken dialogue simulation.
\newblock In {\em 2006 IEEE International Conference on Multimedia and Expo}.
  IEEE, 2006.

\bibitem{pietquin2006probabilistic}
Olivier Pietquin and Thierry Dutoit.
\newblock A probabilistic framework for dialog simulation and optimal strategy
  learning.
\newblock {\em IEEE Transactions on Audio, Speech, and Language Processing},
  2006.

\bibitem{pietquin2013survey}
Olivier Pietquin and Helen Hastie.
\newblock A survey on metrics for the evaluation of user simulations.
\newblock {\em The knowledge engineering review}, 2013.

\bibitem{schatzmann2007error}
Jost Schatzmann, Blaise Thomson, and Steve Young.
\newblock Error simulation for training statistical dialogue systems.
\newblock In {\em IEEE Workshop on Automatic Speech Recognition \&
  Understanding}, 2007.

\bibitem{schatzmann2006survey}
Jost Schatzmann, Karl Weilhammer, Matt Stuttle, and Steve Young.
\newblock A survey of statistical user simulation techniques for
  reinforcement-learning of dialogue management strategies.
\newblock {\em The knowledge engineering review}, 2006.

\bibitem{schatzmann2009hidden}
Jost Schatzmann and Steve Young.
\newblock The hidden agenda user simulation model.
\newblock {\em IEEE transactions on audio, speech, and language processing},
  17(4):733--747, 2009.

\bibitem{schaul2015prioritized}
Tom Schaul, John Quan, Ioannis Antonoglou, and David Silver.
\newblock Prioritized experience replay.
\newblock {\em arXiv:1511.05952}, 2015.

\bibitem{scheffler2002automatic}
Konrad Scheffler and Steve Young.
\newblock Automatic learning of dialogue strategy using dialogue simulation and
  reinforcement learning.
\newblock In {\em Proceedings of the second international conference on Human
  Language Technology Research}. Morgan Kaufmann Publishers Inc., 2002.

\bibitem{scheffler2003automatic}
Konrad~Haarhoff Scheffler.
\newblock {\em Automatic design of spoken dialogue systems}.
\newblock PhD thesis, University of Cambridge, 2003.

\bibitem{su2016continuously}
Pei-Hao Su, Milica Gasic, Nikola Mrksic, Lina Rojas-Barahona, Stefan Ultes,
  David Vandyke, Tsung-Hsien Wen, and Steve Young.
\newblock Continuously learning neural dialogue management.
\newblock {\em arXiv:1606.02689}, 2016.

\bibitem{sutskever2014sequence}
Ilya Sutskever, Oriol Vinyals, and Quoc~V Le.
\newblock Sequence to sequence learning with neural networks.
\newblock In {\em NIPS}, 2014.

\bibitem{wen2016snapshot}
Tsung-Hsien Wen, Milica Ga{\v{s}}i{\'c}, Nikola Mrk{\v{s}}i{\'c}, Lina~M.
  Rojas-Barahona, Pei-Hao Su, Stefan Ultes, David Vandyke, and Steve Young.
\newblock Conditional generation and snapshot learning in neural dialogue
  systems.
\newblock {\em EMNLP}, 2016.

\bibitem{wen2015semantically}
Tsung-Hsien Wen, Milica Ga{\v{s}}i{\'c}, Nikola Mrk{\v{s}}i{\'c}, Pei-Hao Su,
  David Vandyke, and Steve Young.
\newblock Semantically conditioned lstm-based natural language generation for
  spoken dialogue systems.
\newblock {\em EMNLP}, 2015.

\bibitem{williams2016end}
Jason~D Williams and Geoffrey Zweig.
\newblock End-to-end lstm-based dialog control optimized with supervised and
  reinforcement learning.
\newblock {\em arXiv:1606.01269}, 2016.

\bibitem{zhao2016towards}
Tiancheng Zhao and Maxine Eskenazi.
\newblock Towards end-to-end learning for dialog state tracking and management
  using deep reinforcement learning.
\newblock {\em arXiv:1606.02560}, 2016.

\end{thebibliography}

\appendix
\label{sec:appendix-sample}

\section{Recipes}
This framework provides you a way to develop and compare different algorithms/models (i.e., agents in the dialogue setting). The dialogue system consists of two parts: agent and user simulator. Here, we walk through some examples to show how to build and plug in your own agents and user simulators.

\subsection{How to build your own agent?}

For all the agents, they are inherited from the \textcolor{blue}{Agent} class (\textcolor{pythonfile}{agent.py}) which provides some common interfaces for users to implement their agents. In the \textcolor{pythonfile}{agent\_baseline.py} file, five basic rule-based agents are implemented:

\begin{itemize}
\item \textcolor{blue}{\emph{InformAgent}} informs all the slots one by one in every turn; it cannot request any information/slot.
\item \textcolor{blue}{\emph{RequestAllAgent}} requests all the slots one by one in every turn; it cannot inform any information/slot.
\item \textcolor{blue}{\emph{RandomAgent}} requests any random request in every turn; it cannot inform any information/slot.
\item \textcolor{blue}{\emph{EchoAgent}} informs the slot in the request slots of last user action; it cannot request any information/slot.
\item \textcolor{blue}{\emph{RequestBasicsAgent}} requests all basic slots in a subset one by one, then chooses inform(taskcomplete) at the last turn; it cannot inform any information/slot.
\end{itemize}

All the agents just re-implement two functions: \textcolor{blue}{initialize\_episode} and \textcolor{blue}{state\_to\_action}. Here \textcolor{blue}{state\_to\_action} function makes no assumption about the structure of the agent, it is an interface to implement the mapping from state to action, which is the core part in the agent. Here is an example of \textcolor{blue}{\emph{RequestBasicsAgent}}:

\begin{lstlisting}[language=Python, caption=RequestBasicsAgent]
class RequestBasicsAgent(Agent):
    """ A simple agent to test the system. This agent should simply request all the basic slots and then issue: thanks(). """
    
    def initialize_episode(self):
        self.state = {}
        self.state['diaact'] = 'UNK'
        self.state['inform_slots'] = {}
        self.state['request_slots'] = {}
        self.state['turn'] = -1
        self.current_slot_id = 0
        self.request_set = ['moviename', 'starttime', 'city', 'date', 'theater', 'numberofpeople']
        self.phase = 0

    def state_to_action(self, state):
        """ Run current policy on state and produce an action """
        
        self.state['turn'] += 2
        if self.current_slot_id < len(self.request_set):
            slot = self.request_set[self.current_slot_id]
            self.current_slot_id += 1

            act_slot_response = {}
            act_slot_response['diaact'] = "request"
            act_slot_response['inform_slots'] = {}
            act_slot_response['request_slots'] = {slot: "UNK"}
            act_slot_response['turn'] = self.state['turn']
        elif self.phase == 0:
            act_slot_response = {'diaact': "inform", 'inform_slots': {'taskcomplete': "PLACEHOLDER"}, 'request_slots': {}, 'turn':self.state['turn']}
            self.phase += 1
        elif self.phase == 1:
            act_slot_response = {'diaact': "thanks", 'inform_slots': {}, 'request_slots': {}, 'turn': self.state['turn']}
        else:
            raise Exception("THIS SHOULD NOT BE POSSIBLE (AGENT CALLED IN UNANTICIPATED WAY)")
        return {'act_slot_response': act_slot_response, 'act_slot_value_response': None}

\end{lstlisting}

All the above rule-based agents can support only either inform or request action, here you can practice to implement a sophisticated rule-based agent which can support multiple actions, including inform, request, confirm\_question, confirm\_answer, deny etc.  

\textcolor{pythonfile}{agent\_dqn.py} provides a RL agent (\textcolor{codegreen}{agt=9}), which wraps a DQN model. Besides the two above functions, there are two major functions in the RL agent: \textcolor{blue}{run\_policy} and \textcolor{blue}{train}. \textcolor{blue}{run\_policy} implements an $\epsilon$-greedy policy, and \textcolor{blue}{train} calls the batch training function of DQN.

\begin{lstlisting}[language=Python, caption=Two major functions for RL agent]
class AgentDQN(Agent):
    def run_policy(self, representation):
        """ epsilon-greedy policy """
        
        if random.random() < self.epsilon:
            return random.randint(0, self.num_actions - 1)
        else:
            if self.warm_start == 1:
                if len(self.experience_replay_pool) > self.experience_replay_pool_size:
                    self.warm_start = 2
                return self.rule_policy()
            else:
                return self.dqn.predict(representation, {}, predict_model=True)
    
    def train(self, batch_size=1, num_batches=100):
        """ Train DQN with experience replay """
        
        for iter_batch in range(num_batches):
            self.cur_bellman_err = 0
            for iter in range(len(self.experience_replay_pool)/(batch_size)):
                batch = [random.choice(self.experience_replay_pool) for i in xrange(batch_size)]
                batch_struct = self.dqn.singleBatch(batch, {'gamma': self.gamma}, self.clone_dqn)

\end{lstlisting}

\textcolor{pythonfile}{agent\_cmd.py} provides a command line agent (\textcolor{codegreen}{agt=0}), which you as an agent can interact with the user simulator. The command line agent supports two types of input: natural language (\textcolor{codegreen}{cmd\_input\_mode=0}) and dialog act(\textcolor{codegreen}{cmd\_input\_mode=1}). Listing ~\ref{lst:NL_cmd} shows an example of command line agent interacting with the user simulator via the natural language; Listing ~\ref{lst:diaact_cmd} shows an example of command line agent interacting with the user simulator via dialog act form. Note:

\begin{itemize}
\item When the last user turn is a \textit{request} action, the system will show a line of suggested available answers in the database for the agent, like the turn 0 in the Listing ~\ref{lst:diaact_cmd} . Both rule-based agents and RL agent, they will answer the user with the slot values from the database. Here a special case for command line agent is, human (as command line agent) might type any random answer to user's request, when the typed answer is not in the database, the state tracker will correct it, and force the agent to use the values from the database in the agent response. For example, in turn 1 of the Listing ~\ref{lst:diaact_cmd} , if you input \textit{inform(theater=amc pacific)}, the actual answer received by the user is  \textit{inform(theater=carmike summit 16)}, because \textit{amc pacific} doesn't exist in the database, to avoid this wired behavior that agent informs the user a unavailable value, we restrict the agent to use the values from the suggested list.
\item The last second turn of agent is usually an \textit{inform(taskcomplete)} in dialog act form or something like ``\textit{Okay, your tickets are booked.}'' in natural language, which is to inform the user simulator that the agent nearly completes the task, and is ready to book the movie tickets.
\item To end a conversation, the last turn of the agent is usually a \textit{thanks()} in dialog act form or ``\textit{thanks}'' in natural language.
\end{itemize}

\begin{lstlisting}[language=XML, label={lst:NL_cmd}, caption=An example of command line agent interacting with user simulator with natural language]
python run.py --agt 0 --usr 1 --max_turn 40 --episodes 150 --movie_kb_path .\deep_dialog\data\movie_kb.1k.json --goal_file_path .\deep_dialog\data\user_goals_first_turn_template.part.movie.v1.p --intent_err_prob 0.00 --slot_err_prob 0.00 --episodes 500 --act_level 0 --run_mode 0 --cmd_input_mode 0

New episode, user goal:
{
  "request_slots": {
    "ticket": "UNK"
  },
  "diaact": "request",
  "inform_slots": {
    "city": "seattle",
    "numberofpeople": "2",
    "theater": "amc pacific place 11 theater",
    "starttime": "9:00 pm",
    "date": "tomorrow",
    "moviename": "deadpool"
  }
}
Turn 0 usr: Can I buy tickets for deadpool at seattle?
Turn 1 sys: Which city do you want to buy the ticket?
Turn 2 usr: I want to watch at seattle.
Turn 3 sys: Which theater do you want?
Turn 4 usr: I want to watch at amc pacific place 11 theater.
Turn 5 sys: What date would you like?
Turn 6 usr: I want to set it up tomorrow
Turn 7 sys: And what start time do you like?
Turn 8 usr: I want to watch at 9:00 pm.
Turn 9 sys: How many tickets do you need?
Turn 10 usr: I want 2 tickets please!
Turn 11 sys: Okay, your tickets were booked.
Turn 12 usr: Thank you
Turn 13 sys: thanks
Successful Dialog!
\end{lstlisting}

\begin{lstlisting}[language=XML, label={lst:diaact_cmd}, caption=An example of command line agent interacting with user simulator with Dialog Act]
python run.py --agt 0 --usr 1 --max_turn 40 --episodes 150 --movie_kb_path .\deep_dialog\data\movie_kb.1k.json --goal_file_path .\deep_dialog\data\user_goals_first_turn_template.part.movie.v1.p --intent_err_prob 0.00 --slot_err_prob 0.00 --episodes 500 --act_level 0 --run_mode 0 --cmd_input_mode 1

New episode, user goal:
{
  "request_slots": {
    "ticket": "UNK",
    "theater": "UNK"
  },
  "diaact": "request",
  "inform_slots": {
    "city": "birmingham",
    "numberofpeople": "2",
    "state": "al",
    "starttime": "4 pm",
    "date": "today",
    "moviename": "deadpool"
  }
}
Turn 0 usr: Which theater will play the deadpool at 4 pm?
(Suggested Values: {'theater': ['carmike summit 16']})
Turn 1 sys: inform(theater=carmike summit 16)
Turn 2 usr: I need tickets at al.
Turn 3 sys: request(numberofpeople)
Turn 4 usr: I want 2 tickets please!
Turn 5 sys: request(city)
Turn 6 usr: I want to watch at birmingham.
Turn 7 sys: request(starttime)
Turn 8 usr: I want to watch at 4 pm.
Turn 9 sys: request(date)
Turn 10 usr: I want to set it up today
Turn 11 sys: inform(taskcomplete)
Turn 12 usr: Thank you
Turn 13 sys: thanks()
Successful Dialog!
\end{lstlisting}

\subsection{How to build your own user simulator?}
Similarly, there is one user simulator class (\textcolor{pythonfile}{usersim.py}) which provides a few common interfaces for users to implement their user simulators. All the user simulators are inherited from this class, they should re-implement these two functions: \textcolor{blue}{initialize\_episode} and \textcolor{blue}{next}. The \textcolor{pythonfile}{usersim\_rule.py} file implements a rule-based user simulator. Here the \textcolor{blue}{next} function implements all the rules and mechanism to issue the next user action based on the last agent action. Here is the example of \textcolor{pythonfile}{usersim\_rule.py}:

\begin{lstlisting}[language=Python, caption=User Simulator Rules]
def next(self, system_action):
        """ Generate next User Action based on last System Action """
        
        self.state['turn'] += 2
        self.episode_over = False
        self.dialog_status = dialog_config.NO_OUTCOME_YET
        
        sys_act = system_action['diaact']
        
        if (self.max_turn > 0 and self.state['turn'] > self.max_turn):
            self.dialog_status = dialog_config.FAILED_DIALOG
            self.episode_over = True
            self.state['diaact'] = "closing"
        else:
            self.state['history_slots'].update(self.state['inform_slots'])
            self.state['inform_slots'].clear()

            if sys_act == "inform":
                self.response_inform(system_action)
            elif sys_act == "multiple_choice":
                self.response_multiple_choice(system_action)
            elif sys_act == "request":
                self.response_request(system_action) 
            elif sys_act == "thanks":
                self.response_thanks(system_action)
            elif sys_act == "confirm_answer":
                self.response_confirm_answer(system_action)
            elif sys_act == "closing":
                self.episode_over = True
                self.state['diaact'] = "thanks"

        self.corrupt(self.state)
        
        response_action = {}
        response_action['diaact'] = self.state['diaact']
        response_action['inform_slots'] = self.state['inform_slots']
        response_action['request_slots'] = self.state['request_slots']
        response_action['turn'] = self.state['turn']
        response_action['nl'] = ""
        
        # add NL to dia_act
        self.add_nl_to_action(response_action)                       
        return response_action, self.episode_over, self.dialog_status

\end{lstlisting}

\section{Training Details}

To train a RL agent, you can either start with some rule policy experience tuples to initialize the experience replay buffer pool or start with an empty experience replay buffer pool. We recommend to use some rule or supervised policy to initialize the experience replay buffer pool, many work~\cite{williams2016end,su2016continuously,zhao2016towards,lipton2016efficient} have demonstrated the benefits of such strategy as a good initialization to speed up the RL training. Here, we use a very simple rule-based policy to initialize the experience replay buffer pool. 

The RL agent is a DQN network. In the training, we use the $\epsilon$-greedy policy and a dynamic experience replay buffer pool. The size of experience replay buffer pool is dynamic changing. One important trick of DQN is to introduce the target network, which is updated slowly and used to compute the target value in a short period.

\begin{figure}
\centering
\includegraphics[width=1.0\columnwidth]{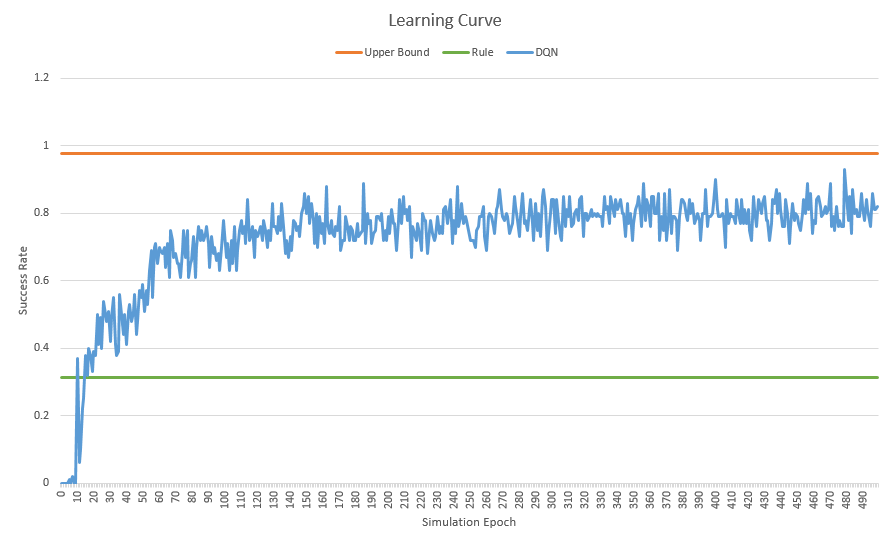}
\caption{Learning curve for policy training, without NLU and NLG: Green line is a rule agent which we employ to initialize the experience replay buffer pool; the blue line is the learning curve for the RL agent; orange line is the optimal upper bound, which is computed by the ratio of the number of reachable user goals in the database of the agent to the total number of user goals.}
\label{fig:rl_learning_curve}
\end{figure}

\begin{figure}
\centering
\includegraphics[width=1.0\columnwidth]{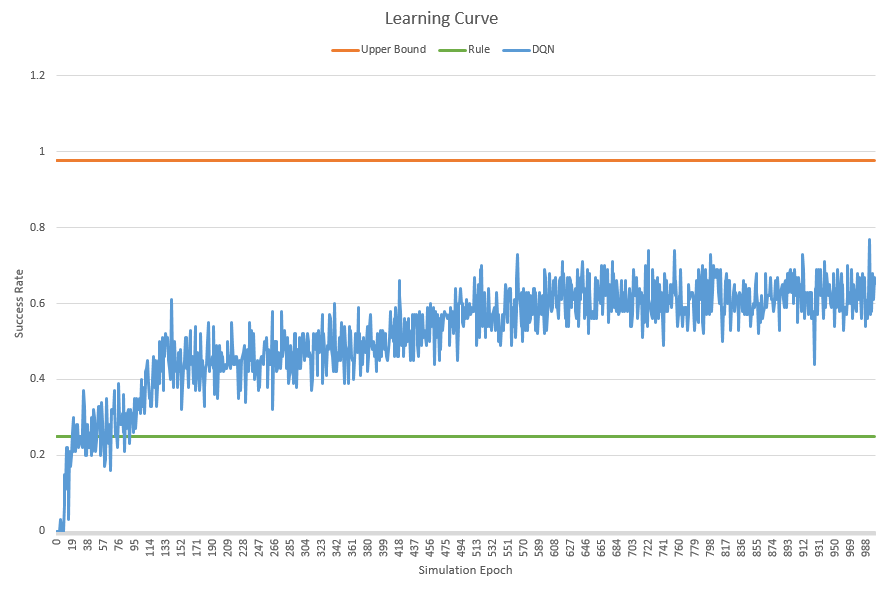}
\caption{Learning curve for the end-to-end policy training, with NLU and NLG: Green line is a rule agent which we employ to initialize the experience replay buffer pool; the blue line is the learning curve for the RL agent; orange line is the optimal upper bound, which is computed by the ratio of the number of reachable user goals in the database of the agent to the total number of user goals.}
\label{fig:rl_learning_curve_e2e}
\end{figure}

The training procedure goes like this way: at each simulation epoch, we simulate $N$ dialogues and add these state transition tuples ($s_t, a_t, r_t, s_{t+1}$) into experience replay buffer pool, train and update the current DQN network. In one simulation epoch, the current DQN network will be updated multiple times, depending on the batch size and the current size of experience replay buffer, at the end of simulation epoch, the target network will be replaced by the current DQN network, the target DQN network is only updated for once in one simulation epoch. The experience replay strategy is critic for the training~\cite{schaul2015prioritized}. Our experience reply buffer update strategy is as follows: at the beginning, we will accumulate all the experience tuples from the simulation and flush the experience reply buffer pool till the current RL agent reaches a success rate threshold (\ie\textit{success\_rate\_threshold} = 0.30), then use the experience tuples from the current RL agent to re-fill the buffer. The intuition behind is the initial performance of the DQN is not good to generate enough good experience replay tuples, thus we do not flush the experience replay pool till the current RL agent can reach a certain success rate which we think it is good, for example, the performance of a rule-based agent. Then in the following training process, at every simulation epoch, we estimate the success rate of the current DQN agent, if the current DQN agent is better enough (\ie better than the target network), the experience replay buffer poll will be flushed and re-filled. Figure \ref{fig:rl_learning_curve} shows a learning curve for RL agent without NLU and NLG, Figure \ref{fig:rl_learning_curve_e2e} is a learning curve for RL agent with NLU and NLG, it takes longer time to train the RL agent to adapt the errors and noise from NLU and NLG.

\section{Sample Dialogues}
\subsection{Task-Completion Bot}
Table \ref{tab:tcp_sample} shows one success and one failure dialogue examples generated by the rule-based agent and RL agent interacting with user simulator in the movie-booking domain. To be informative, we also explicitly show the user goal at the head of the dialogue, but the agent knows nothing about the user goal, its goal is to help the user to accomplish this goal and book the right movie tickets.

\begin{table*}[htbp!]
\small
\centering
\caption{Two sample dialogues generated by rule-based agent and RL agent with user simulator: Left column shows both rule and RL agents succeed; Right column shows that rule-based agent fails, while RL agent succeeds.}
\label{tab:tcp_sample}
\begin{tabular}{l|l}
\hline
\begin{tabular}[c]{@{}l@{}}
New episode, user goal:\\
\{\\
\-\hspace{3mm} "request\_slots": \{	\\
\-\hspace{6mm}    "ticket": "UNK"	\\
\-\hspace{3mm}  \},	\\
\-\hspace{3mm}  "diaact": "request",	\\
\-\hspace{3mm}  "inform\_slots": \{	\\
\-\hspace{6mm}   "city": "seattle",	\\
\-\hspace{6mm}    "numberofpeople": "2",	\\
\-\hspace{6mm}    "theater": "regal meridian 16",	\\
\-\hspace{6mm}    "starttime": "9:25 pm",	\\
\-\hspace{6mm}    "date": "tomorrow",	\\
\-\hspace{6mm}    "moviename": "zoolander 2"	\\
\-\hspace{3mm}  \}\\
\}\\
\end{tabular}  & \begin{tabular}[c]{@{}l@{}}
New episode, user goal:\\
\{\\
\-\hspace{3mm} "request\_slots": \{	\\
\-\hspace{6mm}    "ticket": "UNK"	\\
\-\hspace{6mm}    "theater": "UNK"	\\
\-\hspace{6mm}    "starttime": "UNK"	\\
\-\hspace{3mm}  \},	\\
\-\hspace{3mm}  "diaact": "request",	\\
\-\hspace{3mm}  "inform\_slots": \{	\\
\-\hspace{6mm}    "numberofpeople": "3",	\\
\-\hspace{6mm}    "date": "tomorrow",	\\
\-\hspace{6mm}    "moviename": "10 cloverfield lane"	\\
\-\hspace{3mm}  \}\\
\}\\
\-\hspace{3mm}
\end{tabular}                \\
\hline
\textbf{Rule-based agent} & \textbf{Rule-based Agent} \\
\hline

\underline{User}: Can I get some tickets for zoolander 2?  & \underline{User}: What is the start time for 10 cloverfield lane?  \\
\underline{Agent}: What movie are you interested in? & \underline{Agent}: What movie are you interested in?  \\
\underline{User}: I want to watch zoolander 2.  & \underline{User}: I want to watch at 10 cloverfield lane.  \\
\underline{Agent}: What time would you like to see it? & \underline{Agent}:What time would you like to see it? \\
\underline{User}: I want to watch at 9:25 pm. & \underline{User}: hat time is it playing tomorrow? I need 3 tickets. \\
\underline{Agent}: What city you would like? & What city you would like?  \\
\underline{User}: I want to watch at seattle. & \underline{User}:I do not care.   \\
\underline{Agent}: What date would you like to watch it? &  What date would you like to watch it? \\
\underline{User}: I want to set it up tomorrow. &  \underline{User}: I want to set it up tomorrow.  \\   
\underline{Agent}: Which theater would you like?   &    \underline{Agent}: Which theater would you like?    \\
\underline{User}: I want to watch at regal meridian 16.	&    \underline{User}: Which theater is available?      \\
\underline{Agent}: How many tickets do you need?  &  \underline{Agent}: How many tickets do you need?   \\
\underline{User}: I want 2 tickets please! &  \underline{User}: I want 3 tickets please! \\ 
\begin{tabular}[c]{@{}l@{}}\underline{Agent}: Great - I was able to purchase 2 tickets for \\ you to see zoolander 2 tomorrow at regal meridian 16 \\ theater in seattle at 9:25 pm. \end{tabular}  &  \begin{tabular}[c]{@{}l@{}}\underline{Agent}: Okay, I have booked 3 tickets for 10 cloverfield \\ lane tomorrow. \end{tabular}     \\
\underline{User}: Thank you. & \underline{User}: Thank you. \\ 
\underline{Agent}: Thank you. & \underline{Agent}: Thank you. \\ 
\hline
\textbf{Success} & \begin{tabular}[c]{@{}l@{}} \textbf{Failure:} Agent does not answer all the questions \\ (starttime, theater) from user.\end{tabular}
\\ \hline

\textbf{RL agent} & \textbf{RL Agent} \\
\hline

\underline{User}: Can I get tickets for zoolander 2 tomorrow?  & \begin{tabular}[c]{@{}l@{}}\underline{User}: Which theater can I book 3 tickets for \\ 10 cloverfield lane?  \end{tabular} \\
\underline{Agent}: Which city would you like?    & \underline{Agent}: What time would you like to see it?  \\
\underline{User}: I want to watch at seattle.    & \begin{tabular}[c]{@{}l@{}} \underline{User}: Which theater and start time are available \\ tomorrow?  \end{tabular}\\
\underline{Agent}: How many tickets do you need?  & \underline{Agent}: 11:45am is available.     \\
\underline{User}: I want 2 tickets please!      & \underline{User}: Which theater is available? \\
\underline{Agent}: 9:25 pm is available.       & \underline{Agent}: regal la live stadium 14 is available.  \\
\underline{User}: I want to watch at regal meridian 16.   & \underline{User}: Could you help me to book the tickets?  \\
\begin{tabular}[c]{@{}l@{}}\underline{Agent}: Great - I was able to purchase 2 tickets for \\ you to see zoolander 2 tomorrow at regal meridian 16 \\ theater in seattle at 9:25 pm. \end{tabular}   &   \begin{tabular}[c]{@{}l@{}} \underline{Agent}: Okay - I was able to book 3 tickets for you to \\ see 10 cloverfield lane at regal la live stadium 14 at \\ 11:45am tomorrow. \end{tabular}      \\
\underline{User}: Thank you.    &      \underline{User}:Thank you  \\   
\underline{Agent}: Thank you.   &      \underline{Agent}: Thank you     
\\ \hline
\textbf{Success} & \textbf{Success}
\\ \hline

\end{tabular}
\end{table*}

\subsection{KB-InfoBot}
Table~\ref{tab:sample} shows some sample dialogues between the user simulator and SimpleRL-SoftKB and End2End-RL agents~\cite{dhingra2016end}. Value of the \textit{critic\_rating} slot is a common source of error in the user simulator, and hence all learned policies tend to ask for this value multiple times.

\begin{table*}[htbp!]
\small
\centering
\caption{Sample dialogues between user simulator and SimpleRL-SoftKB and End2End-RL agents. At the end of each dialogue, the agent informs top $5$ results from the KB posterior. User target, if informed, is in bold.}
\label{tab:sample}
\begin{tabular}{l|l}
\hline
\textbf{SimpleRL-SoftKB}  & \textbf{End-to-End}   \\ \hline
\begin{tabular}[c]{@{}l@{}}\underline{User}: i would like to know the moviename whose \\ genre is drama and critic\_rating is 6.3.\end{tabular}   & \begin{tabular}[c]{@{}l@{}}\underline{User}: i would like to know the moviename whose \\ genre is drama and critic\_rating is 6.3.\end{tabular}                \\
\underline{Agent}: \textit{request}(mpaa\_rating)  & \underline{Agent}: \textit{request}(actor) \\
\underline{User}: i cannot remember & \underline{User}: i cannot remember   \\
\underline{Agent}: \textit{request}(actor)  & \underline{Agent}: \textit{request}(director)   \\
\underline{User}: i cannot remember  & \underline{User}: milan cheylov director  \\
\underline{Agent}: \textit{request}(critic\_rating)  & \begin{tabular}[c]{@{}l@{}}\underline{Agent}: \textit{inform}(\textbf{the countess}, hemingway \& gellhorn, \\ adios derek, unstoppable, wagers of sin)\end{tabular}  \\
\underline{User}: i think it is 6.3   &  \\
\underline{Agent}: \textit{request}(critic\_rating)  & \\
\underline{User}: i think it is 6.3  & \\
\begin{tabular}[c]{@{}l@{}}\underline{Agent}: \textit{inform}(kid svensk, hemigway \& gellhorn, \\ ry. ga gotuku, night catches us, tmnt)\end{tabular}    &                                                                                                                   \\ \hline

\begin{tabular}[c]{@{}l@{}}\underline{User}: i am looking for a moviename with mpaa\_rating \\ 6 and director .\end{tabular}                                                     & \begin{tabular}[c]{@{}l@{}}\underline{User}: i am looking for a moviename with mpaa\_rating \\ 6 and critic\_rating 8.7.\end{tabular}                          \\
\underline{Agent}: \textit{request}(actor)                                                                                                                                                & \underline{Agent}: \textit{request}(actor)                                                                                                                              \\
\underline{User}: i cannot remember                                                                                                                                              & \underline{User}: i cannot remember                                                                                                                            \\
\underline{Agent}: \textit{request}(critic\_rating)                                                                                                                                       & \underline{Agent}: \textit{request}(director)                                                                                                                           \\
\underline{User}: i think it is 7.5                                                                                                                                              & \underline{User}: i cannot remember                                                                                                                            \\
\underline{Agent}: \textit{request}(genre)                                                                                                                                                & \underline{Agent}: \textit{request}(critic\_rating)                                                                                                                     \\
\underline{User}: i cannot remember                                                                                                                                              & \underline{User}: i think it is 7.5                                                                                                                            \\
\underline{Agent}: \textit{request}(critic\_rating)                                                                                                                                       & \underline{Agent}: \textit{request}(critic\_rating)                                                                                                                     \\
\underline{User}: i think it is 7.5                                                                                                                                              & \underline{User}: i think it is 7.5                                                                                                                            \\
\underline{Agent}: \textit{request}(critic\_rating)     & \underline{Agent}: \textit{request}(genre)  \\
\underline{User}: i think it is 4.0   & \underline{User}: i cannot remember    \\
\begin{tabular}[c]{@{}l@{}}\underline{Agent}: \textit{inform}(l'ennemi public, open season, \\ \textbf{october road}, fleeting cheating meeting, final draft)\end{tabular}                         & \begin{tabular}[c]{@{}l@{}}\underline{Agent}: \textit{inform}(fleeting cheating meeting, \textbf{october road}, \\ open season, l'ennemi public, journeyman)\end{tabular}        \\ \hline
\begin{tabular}[c]{@{}l@{}}\underline{User}: which moviename with director pg director \\ khan and mpaa\_rating .\end{tabular}                                                   & \begin{tabular}[c]{@{}l@{}}\underline{User}: i am looking for a moviename with director \\ khan critic\_rating 2008 and mpaa\_rating pg.\end{tabular}          \\
\underline{Agent}: \textit{request}(actor)  & \underline{Agent}: \textit{request}(actor)   \\
\underline{User}: i cannot remember                                                                                                                                              & \underline{User}: i cannot remember                                                                                                                            \\
\begin{tabular}[c]{@{}l@{}}\underline{Agent}: inform(heber holiday, husbands and knives, \\ not quite hollywood: the wild, the man who \\ lost his head, \textbf{meet dave})\end{tabular} & \underline{Agent}: \textit{request}(critic\_rating)    \\
& \underline{User}: i cannot remember  \\
& \underline{Agent}: \textit{request}(genre)  \\
& \underline{User}: i cannot remember  \\
& \begin{tabular}[c]{@{}l@{}}\underline{Agent}: \textit{inform}(heber holiday, not quite hollywood: \\ the wild, \textbf{meet dave}, agent crush, mrs mcginty's dead)\end{tabular} \\ \hline
\begin{tabular}[c]{@{}l@{}}\underline{User}: i am looking for a moviename with actor \\ dylan release\_year 2008 and genre horror.\end{tabular}   & \begin{tabular}[c]{@{}l@{}}\underline{User}: i am looking for a moviename with actor \\ dylan release\_year 2008 and genre horror.\end{tabular}    \\
\begin{tabular}[c]{@{}l@{}}\underline{Agent}: \textit{inform}(\textbf{parasomnia}, encarna..o do dem.nio, \\ sense \& sensibility, dead and gone, the broken)\end{tabular}   & \begin{tabular}[c]{@{}l@{}}\underline{Agent}: \textit{inform}(\textbf{parasomnia}, encarna..o do dem.nio, \\ the broken, sense \& sensibility, american teen)\end{tabular}    \\ \hline
\underline{User}: which moviename has natasha as actor?                                                                                                                          & \underline{User}: which moviename has natasha as actor?                                                                                                        \\
\underline{Agent}: \textit{request}(mpaa\_rating)                                                                                                                                         & \underline{Agent}: \textit{request}(critic\_rating)                                                                                                                     \\
\underline{User}: i cannot remember                                                                                                                                              & \underline{User}: i think it is 9.2.                                                                                                                           \\
\underline{Agent}: \textit{request}(genre)                                                                                                                                                & \underline{Agent}: \textit{request}(critic\_rating)                                                                                                                     \\
\underline{User}: music                                                                                                                                                          & \underline{User}: i think it is 8.9.                                                                                                                           \\
\underline{Agent}: \textit{request}(critic\_rating)                                                                                                                                       & \underline{Agent}: \textit{request}(director)                                                                                                                           \\
\underline{User}: 9.3                                                                                                                                                            & \underline{User}: i cannot remember                                                                                                                            \\
\begin{tabular}[c]{@{}l@{}}\underline{Agent}: \textit{inform}(the belgrade phantom, nine inch \\ nails live, zhan. gu, taare zameen par, elvis lives)\end{tabular}                        & \underline{Agent}: \textit{request}(mpaa\_rating)   \\
& \underline{User}: i cannot remember   \\                                                                                               & \underline{Agent}: \textit{request}(actor) \\ & \underline{User}: the actor is natasha alderslade.                                                                                                             \\
                                                                                                                                                                     & \begin{tabular}[c]{@{}l@{}}\underline{Agent}: \textit{inform}(\textbf{mugged}, the rage, wayne's world, \\ arritmia, the broken)\end{tabular}                                    \\ \hline
\end{tabular}
\end{table*}

\end{document}